\pdfoutput=1

\documentclass[11pt]{article}

\usepackage{EMNLP2022}

\usepackage{times}
\usepackage{latexsym}

\usepackage[T1]{fontenc}

\usepackage[utf8]{inputenc}

\usepackage{microtype}

\usepackage{inconsolata}

\usepackage{graphicx}
\usepackage{linguex}

\title{What A Situated Language-Using Agent Must be\\ Able to Do: A Top-Down Analysis}

\author{David Schlangen \\
  Computational Linguistics / Department of Linguistics \\
  University of Potsdam, Germany\\
  \texttt{david.schlangen@uni-potsdam.de}}

\begin{document}
\maketitle
\begin{abstract}
  Even in our increasingly text-intensive times, the primary site of language use is situated, co-present interaction.
  Situated interaction is also the final frontier of Natural Language Processing (NLP),
  where, compared to the area of text processing, little progress has been made in the past decade, and where a myriad of practical applications is waiting to be unlocked.
 While the usual approach in the field is to reach, bottom-up, for the ever next ``adjacent possible'', in this paper I attempt a top-down analysis of what the demands are that unrestricted situated interaction makes on %
the participating agent,
 and suggest ways in which this analysis can structure computational models and research on them.
Specifically, I discuss representational demands (the building up and application of world model, language model, situation model, discourse model, and agent model) and what I call anchoring processes (incremental processing, incremental learning, conversational grounding, multimodal grounding) that bind the agent to the here, now, and us.
\end{abstract}

\section{Introduction}

As \citet{Bisk2020} have noted, 
NLP as a field is slowly working its way towards ever wider ``world scopes'', %
going from modelling corpora to larger collections of text, to collections of text paired with other modalities, to modelling in environments over which the learning agent has some control, currently reaching out to scenarios where other agents need to be modelled as well.
It is interesting to note how curiously backwards this would be as a description of the development of a human language user:
Humans needs to experience other minds before they can ever begin to experience structured textual information. 
As a development strategy, the bottom-up methodology reflected in this ``widening of scopes'' also bears some risks: As \citet{koller:topdown} recently argued with respect to distributional semantics, a bottom-up strategy by design moves from one (relative) success to the next, as the next thing is always the one that is just about possible to do. Without some further guidance, however, this %
limits the perspective and comes with the risk of getting stuck in local optima. This paper is an attempt to provide such guidance for the field of ``embodied social AI'', by pulling together some of what is known in the various scientific areas that deal with human verbal interaction, into an abstract description of modelling desiderata. In that sense, the proposal here may serve as a \emph{conceptual benchmark} against which settings, tasks, datasets, and models can be measured in terms of their coverage, and in terms of the costs of the abstractions they make relative to this general model.\footnote{%
  Note that this way of proceeding is fully compatible with an ``empirical approach'', insofar as that is used to select the best model, and does not aim to determine the goals as well.
}

\begin{figure}[t]
  \centering
  \hspace*{-3ex}
  \includegraphics[width=1.1\linewidth]{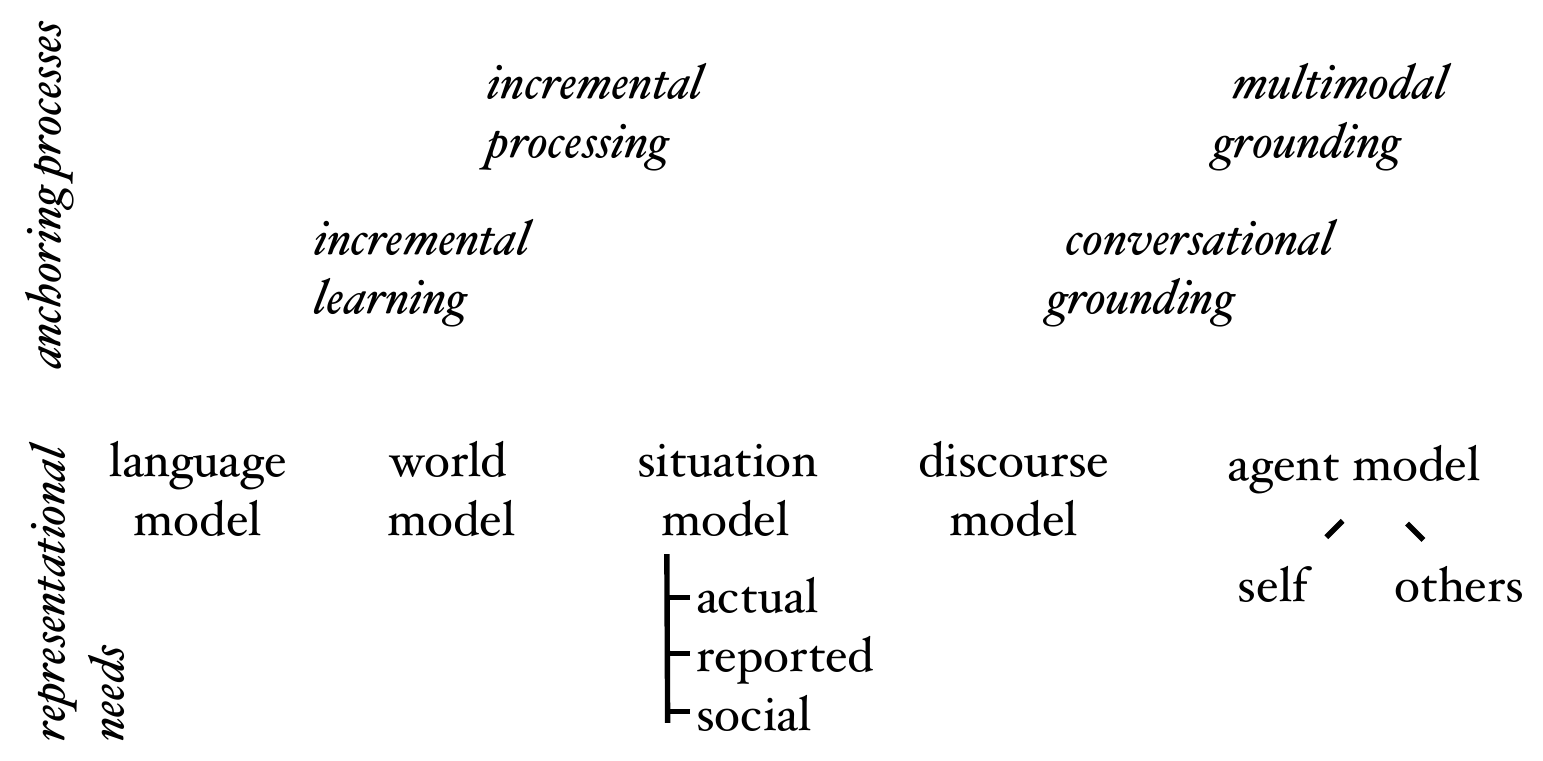}
  \caption{Representational Domains (bottom) and Anchoring Processes (top) Structuring the Situated Agent}
  \label{fig:analysis}
  \vspace*{-2.5ex}
\end{figure}

Figure~\ref{fig:analysis} shows an outline of the proposal, which the remaining sections will unpack: There are representational demands that the situation puts on the agent---that is, the agent needs to bring some knowledge, and track some information (discussed further in Section~\ref{sec:repr}). The processes with which it handles the interaction (and comes to update this knowledge) also are subjects to some demands, stemming from the fact that the interaction partner is free and independent, but similar (Sections~\ref{sec:sitlau}, \ref{sec:anchor}).

It is useful to clarify one thing from the outset: What this paper is \emph{not} trying to do is to make any recommendations as to \emph{how} aspects of this model are to be realised (e.g., using symbolic or distributed representation methods; using particular learning algorithms; building in a certain modularisation or modelling monolithically; or using particular decision making algorithms);
the intended contribution is an analysis of how the %
\emph{phenomenon} of situated language use is conceptually structured on a high level, which can then eventually guide the definition of challenges and selection of methods to meet them.

\section{Situated Interaction} %
\label{sec:sitlau}

Here is a (very) high-level, general characterisation of the
face-to-face interaction situation: It is a \emph{direct, purposeful encounter of free and independent, but similar agents}.
Let us unpack this:
\\ $\bullet$ as \emph{agents}, the participants meet their purposes---and here, specifically, \emph{communicative purposes}---through acting;
\\ $\bullet$ as \emph{free} agents, they cannot be forced, and cannot force the respective other, to do anything, and specifically not to \emph{understand} as intended;
\\ $\bullet$ as \emph{independent} agents, they are individually subject to the same passing of time (while one acts, the other can as well and need not wait); they will also have different histories, including their histories of previous interactions and language use, and will bring different knowledge to the interaction;
\\ $\bullet$ this being a \emph{direct} encounter, the agents must rely on what they can do (produce for the other, receive from the other) with their bodies to create meaning here and now;
\\ $\bullet$ finally, as fundamentally \emph{similar} agents, they can rely on a certain body of shared knowledge and experience, for example in how each parses the shared environment, understands the world, and forms desires, beliefs, and intentions, and, if they are to use language for communication, in how they use language, but where the exact degree of similarity will need to be determined during and through the interaction.
\\ This has consequences: To reach \emph{joint} purposes, the agents need to coordinate, in a process that unfolds continuously in time and which can yield new knowledge, including about how to coordinate, but that also can rest on assumed commonalities.\footnote{%
  This short description places a different focus, but in the broad strokes follows the analysis by \citet{clark:ul}.
}

The next sections will go into the details of what the situation, thus characterised, demands of the agent.

\section{Representational Demands}
\label{sec:repr}

The central means through with agents in situated interaction meet their
purposes is \emph{language} (and a particular one, at that), and hence the agent must come with knowledge of this language, or possess (or represent to itself) what I will call here a \textbf{language model}. 
It is not enough for the agent to be able to produce well-formed strings; rather, the systematic connection to the communicative intentions they express \cite{grice:meaning} must be modelled as well. As these intentions can concern objects in the world, and to the degree that the model of the language can be presumed to be shared, it is via those that the language can count as \emph{grounded} in the sense that has most currency in the NLP community \cite{Chandu2021}.

Examples like those in \ref{ex:worldk} below indicate that \textbf{world knowledge} also factors into the purpose-directed use of language:

\ex. \label{ex:worldk}
  \a. \label{ex:wino}
   I couldn't put the coat into the suitcase because it was too small.
  \b. \label{ex:instruction}
   Put the poster up on the wall. %

   In \ref{ex:wino}, a ``Winograd schema'' \citep{Levesque2012} type sentence, information about expectable relative sizes, and in \ref{ex:instruction}, knowledge about expected outcomes, is needed to interpret the utterance.\footnote{%
 But see for example \citet{Pust:GL,murphy2010lexical} for the notorious difficulties separating linguistic, and in particular lexical knowledge from such more general knowledge.
} 
Again, underlying the communicative use of this knowledge is an assumption that it is shared.

While subject to possible updates, as we will see, these types of knowledge can be seen as something that the agent brings into the situation. But the situation itself must be understood by the agents, in order to interact in it.
The proposed schema %
splits the \textbf{situation model} into three sub-types: A model of the \emph{actual situation} in which the interaction is happening, which would provide not only referents for ``poster'' and ``wall'' in a situation in which \ref{ex:instruction} is used, but also potential likely referent for the implicit instrument of the requested action (e.g., perhaps there is a roll of duct tape visible, or a collection of pushpins). For this to work, there is an underlying assumption, which is that the situation will be mostly parsed similarly by the agents, so that it can form the shared basis for assumed mutual knowledge \cite{clark:ul}; repair processes accounting for violations of this assumption will be discussed below.

The discourse of the agents does not always have to be about the actual situation, however. The building up of model of the \emph{reported situation} \cite{van1983strategies}, together with world knowledge about the consequences of entering a room, can explain the licensing of the contrast (indicating surprisal) in the following example:

\ex. \label{ex:continuity}
I saw two people enter the room, but when I followed, the room was empty.

Lastly, the \emph{social situation} also bears on linguistic material: Relative social status, for example, is grammaticalised in many languages \citep{bender2013linguistic}; and even more generally, the simple fact of who is and who is not party to an interaction determines which linguistic and other behaviour is appropriate \citep{goffman1981forms,bohus-horvitz-09:multiparty}.

Next, the \textbf{discourse model}, required to keep track of antecedents of anaphoric acts and, more generally, for the determination of \emph{coherence}. In \ref{ex:dm}, for example, the anaphoric elements \emph{no} (as negating the proposition contained in A's question) and \emph{he} can only be resolved under the assumption that they realise an \emph{answer} (to A's question) and an \emph{explanation} (for the answer), respectively. (See, \emph{inter alia}, \citet{KnR:DtL,ashlasc:sdrtbook,ginz:is}.)

\ex. \label{ex:dm}
  A: Is John coming to the party?\\
  B: No, he's busy.

Finally, there is a large body of work elucidating the role of the \textbf{agent model} (representing their \emph{beliefs, desires, and intentions}) in interpreting discourse \cite{cmp:intcom}. To again give just one illustrating example, in \ref{ex:coffee}, A must know something about B's likely desires and intentions (to stay awake, or not stay awake) to make sense of their reply.

\ex. \label{ex:coffee}
A: Do you want some coffee?\\
B: It's late.

\section{Anchoring Processes}
\label{sec:anchor}

\begin{figure}[ht]
  \centering
  \hspace*{-3ex}
  \includegraphics[width=1\linewidth]{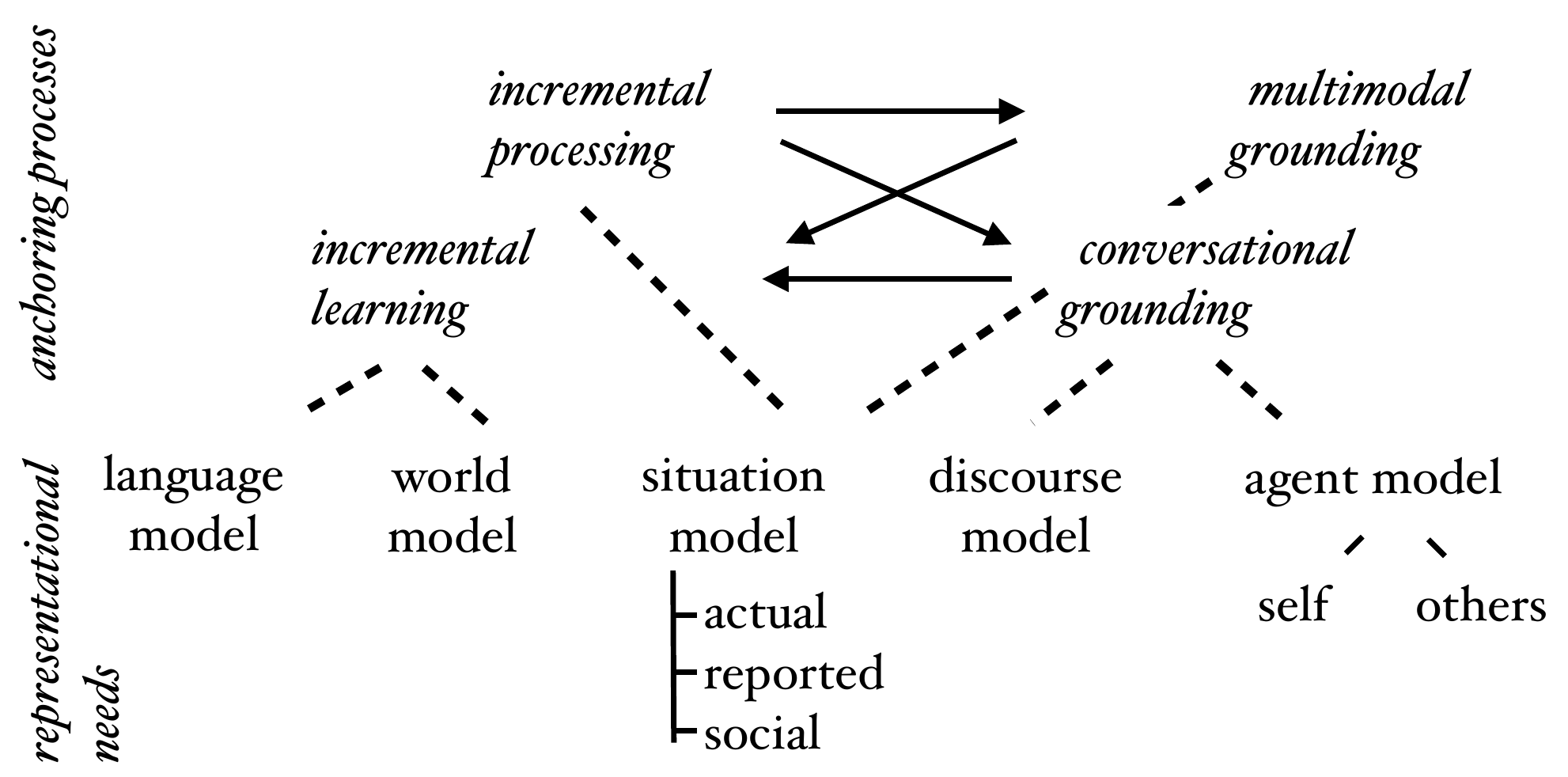}
  \caption{Representations and Processes. Arrows Denote the \emph{enables} Relation, Dotted Lines the \emph{updates} Relation.}
  \label{fig:rels}
  \vspace*{-3ex}
\end{figure}

Even if not often taken in its full breadth, \emph{that} the knowledge described above plays a role in situated interaction is presumably not very controversial. The focus of this section is on something that is less often dealt with and brought together, namely, the processes through which the knowledge is applied and updated. 

The fact that the agents are \emph{independent} and hence not extrinsically temporally coordinated argues for \textbf{incremental processing}, that is, an updating of situation, discourse and agent models that is continual to the observation of the other agent's actions as well as to the agent's own production---this is turn then makes possible the \emph{achievement} of coordination, for example in successful turn-taking \cite{schlaska:agmo}.

Only this processing regime then makes available certain devices used in \textbf{conversational grounding} \citep{clarkbrennan:grounding,clark:ul}---the process of coordination on what can count as shared knowledge (with respect to either of the models discussed above, most immediately the discourse model)---namely the use of overlapping signals such as ``back-channel'' behaviours  like ``uh-hu'' or nodding \cite{yngve:edge}. When understanding is not reached immediately, \emph{repair} can be initiated through clarifications and corrections \cite{hayashi2013conversational}. In the framework of section~\ref{sec:sitlau}, this can be understood as the mending and making true of initially overoptimistic assumptions on what was shared.

In some cases, the outcome of the repair process can lead to ``localised'' learning through the establishment of precedents \cite{brenclark:conpact}, such as for example particular idiosyncratic ways of referring to some object; however, it can also, just like \emph{direct teaching}, lead to a longer term update of language model (e.g., learning new terms) or world model (e.g., learning new facts), in a process of \textbf{incremental learning} \cite{hoppitt:sociallearning}.

Lastly, the multimodal nature of situated interaction \cite{Holler2019} is accounted for by processes of \textbf{multimodal grounding}, which integrate meaning-making devices such as deictic and iconic gestures \cite{Sowa2003,kenningtonetal:sigdial13} and facial expressions \cite{poggipela:facperf}. I will also subsume under this header the process of resolving references into the situational context \cite{royreit:langworld,sieschla:vispento} by performing the categorisations denoted by the expressions.

To summarise the preceding two sections, Figure~\ref{fig:rels} again shows the elements of the analysis discussed here, this time with interrelations added.

\section{Where We Are, And Where To Go} %
\label{sec:where}
\vspace*{-.2cm}

It should be clear that from the perspective of the analysis detailed above, even the NLP tasks that most seem like they are related to situated interaction are severly limited. Here is not the space for a detailed catalogisation, but we can look at a few examples. ``Visual dialog'' \cite{visdial}, the task of generating a reply to a question about an image, within the context of preceding questions and answers, requires a limited discourse model (the previously asked questions and answer may introduce discourse referents), and a limited form of situation model and multimodal grounding (of the target image), but the strict role asymmetry (questioner and answerer) precludes any need for agent modelling and conversational grounding; its strict turn-by-turn nature and the use of written language abstract away from the questioning agent as independent agent and put control unilaterally in the hand of the model. ``Embodied QA'' \citep{dasetal:eqa} and ``language \& vision navigation'' \citep{Anderson2018}, are tasks where in response to a language stimulus actions (in a simulator environment) need to be performed. Hence, these tasks require a more dynamic situation model, but other than that, are not fundamentally different from the visual dialog task (and in fact take away again what little that task requires in terms of discourse model). 

Now, tackling a problem by focussing on its parts is a valid strategy, but only if in doing so the whole is kept in mind. In the cases cited above, it seems fair to say that the formulation of the task was driven more by the available modelling method: They basically are tasks that lend themselves to a formulation as sequence-to-sequence problem, and as such are more about transducing the semantics of the stimulus language than they are about situated \emph{interaction} (or interaction at all). 

More recently, tasks have been proposed that put more stress on the conversational grounding aspects mentioned above \cite{ilieatal:meetup19,Udagawa2019,bara-etal-2021-mindcraft}. This is a good start, but in order to systematise these efforts, what is missing is a clearer picture of how the task setting (environment, interaction mode, etc.) determines what a task can even test, and how close it will come to the fuller picture sketched above. Should it turn out that for the richest settings, real interaction with capable language users is required, then ways will have to be found to enable that, and to overcome the batch learning mode that current models are bound to.

\section{Related Work}
\label{sec:relwo}
\vspace*{-.2cm}

That various kinds of knowledge and update processes are required to model conversational agents is not a new insight.
The grandparent of any of those attempts, Winograd's \citeyear{Winograd:shdrlu} SHDRLU already made a distinction between language model (in the form of parsing procedures) and situation model, as did the later textbook presentation by \citet{allen:nlu}. A distinction between conversation situation and reported situation was made by \citet{barwiseperry:sitatt}; \citet{bratman:intplan} and \citet{wooldridge:bda} stressed the importance of modelling agents in terms of their beliefs, desires, and intentions. \citet{allenetal:archmorereal} were among the first to point out the need for incremental processing. And to conclude this---almost absurdly selective---tour through what is a massive body of work, \citet{traumlarsson:ISbook} and \citet{staffan:thesis} described a representational system that elegantly interfaced discourse modelling, conversational grounding, and agent modelling. It is not, I want to claim here, that the \emph{analyses} from these papers were wrong; in the light of more recently available methods, what is likely the case is that the realisation of representational demands through manually constructed representation formats and formalism restricted these models, and that this is what our more recent methods can help us overcome.\footnote{%
  See \cite{beyond-single} for a very recent, even more wide-ranging argument for the value of looking at situated interaction.
}

\vspace*{-.2cm}
\section{Conclusions}
\label{sec:conc}
\vspace*{-.2cm}

I have argued for a particular analysis of the task of participating in situated interaction, drawing on various literatures. If NLP wants to advance on this phenomenon, I contend, it needs to start to take its complexity seriously, and devise methods and testbeds for tackling it, rather than only invent tasks that fit the available methods.

\bibliography{/Users/das/work/projects/MyDocuments/BibTeX/all-lit.bib}
\bibliographystyle{acl_natbib}

\end{document}